\documentclass[dvips,11pt,a4paper]{article}
\usepackage[hyperref]{acl2017}
\usepackage{ccg}
\usepackage{times}
\usepackage{latexsym}
\usepackage{amsmath}
\usepackage{amssymb}
\usepackage{bm}
\usepackage{graphicx}
\usepackage{tikz-dependency}
\usepackage{subfig}
\usepackage{multirow}
\newcommand{\argmax}{\mathop{\rm arg~max}\limits}

\usepackage{url}

\aclfinalcopy 
\def\aclpaperid{440} 

\title{A* CCG Parsing with a Supertag and Dependency Factored Model}

\author{Masashi Yoshikawa \and Hiroshi Noji \and Yuji Matsumoto \\
Graduate School of Information and Science \\
Nara Institute of Science and Technology \\
8916-5, Takayama, Ikoma, Nara, 630-0192, Japan \\
{\tt \{ masashi.yoshikawa.yh8, noji, matsu \}@is.naist.jp}}

\date{}

\begin{document}
\maketitle
\begin{abstract}
    We propose a new A* CCG parsing
    model in which the probability of a tree is decomposed into
    factors of CCG categories and its syntactic dependencies both defined on bi-directional LSTMs.
    Our factored model allows the precomputation of all probabilities and runs very efficiently,
    while modeling sentence structures explicitly via dependencies.
    Our model achieves the state-of-the-art results on
    English and Japanese CCG parsing.\footnote{
        Our software and the pretrained models are available at:
        \url{https://github.com/masashi-y/depccg}.
    }
\end{abstract}

\section{Introduction}
\label{intro}
Supertagging in lexicalized grammar parsing is known as {\it almost parsing} \cite{bangalore1999supertagging}, in that each supertag is syntactically informative and most ambiguities are resolved once a correct supertag is assigned to every word.
Recently this property is effectively exploited in A* Combinatory Categorial Grammar (CCG; \citet{steedman:syntactic_process}) parsing \cite{lewis-steedman:2014:EMNLP2014,lewis-lee-zettlemoyer:2016:N16-1},
in which the probability of a CCG tree ${\bm y}$ on a sentence ${\bm x}$ of length $N$ is the product of the probabilities of supertags (categories) $c_i$ (locally factored model):
\begin{eqnarray}
    \label{lew}
    P({\bm y} | {\bm x}) = \prod_{i \in [1,N]} P_{tag}(c_i | {\bm x}).
\end{eqnarray}
By not modeling every combinatory rule in a derivation, this formulation enables us to employ efficient A* search (see Section \ref{related}), which finds the most probable supertag sequence that can build a well-formed CCG tree.

\begin{figure}[t]
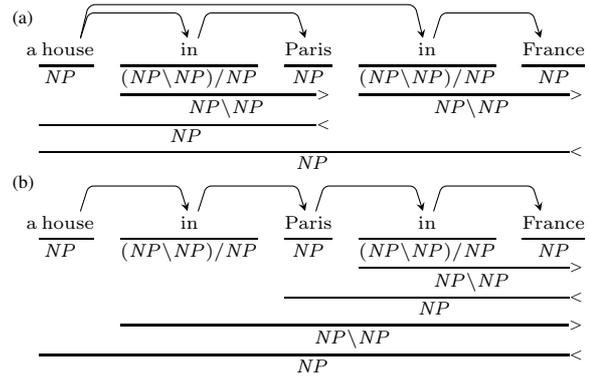

\scriptsize
\centering
\begin{minipage}[t]{1\hsize}
    (a)\hspace{1\hsize}\\[-4ex]
    \subfloat{
\begin{dependency}
   \begin{deptext}[column sep=1.3em, font=\tiny]
       \hspace{5.3em} \& \hspace{5.3em}  \& \hspace{5.3em} \& \hspace{5.3em} \& \hspace{5.3em} \\
   \end{deptext}
   \depedge[hide label, edge height=3ex]{2}{3}{}
   \depedge[hide label, edge height=3ex]{4}{5}{}
   \depedge[hide label, edge height=3ex]{1}{2}{}
   \depedge[hide label, edge height=4ex]{1}{4}{}
\end{dependency}}\\[-4ex]
\subfloat{
\deriv{5}{
{\rm a~house}&{\rm in}& {\rm Paris}& {\rm in}& {\rm France}\\
\uline{1}& \uline{1}& \uline{1}& \uline{1}& \uline{1}\\
\it NP &\it (NP\bs NP)/NP &\it  NP &\it (NP\bs NP)/NP&\it NP\\
& \fapply{2}& \fapply{2}\\
& \mcc{2}{\it NP\bs NP}& \mcc{2}{\it NP\bs NP}\\
\bapply{3}\\
\mcc{3}{\it NP}\\
\bapply{5}\\
\mcc{5}{\it NP}\\
}}
\end{minipage}\\
\begin{minipage}[t]{1\hsize}
    (b)\hspace{1\hsize}\\[-2ex]
\subfloat{
\begin{dependency}
   \begin{deptext}[column sep=1.3em, font=\tiny]
       \hspace{5.3em} \& \hspace{5.3em}  \& \hspace{5.3em} \& \hspace{5.3em} \& \hspace{5.3em} \\
   \end{deptext}
   \depedge[hide label, edge height=3ex]{2}{3}{}
   \depedge[hide label, edge height=3ex]{4}{5}{}
   \depedge[hide label, edge height=3ex]{1}{2}{}
   \depedge[hide label, edge height=3ex]{3}{4}{}
\end{dependency}}\\[-4ex]
\subfloat{
\deriv{5}{
{\rm a~house}&{\rm in}& {\rm Paris}& {\rm in}& {\rm France}\\
\uline{1}& \uline{1}& \uline{1}& \uline{1}& \uline{1}\\
\it NP &\it (NP\bs NP)/NP &\it  NP &\it (NP\bs NP)/NP&\it NP\\
&&& \fapply{2}\\
&&& \mcc{2}{\it NP\bs NP}\\
&&\bapply{3}\\
&& \mcc{3}{\it NP}\\
&\fapply{4}\\
& \mcc{4}{\it NP\bs NP}\\
\bapply{5}\\
\mcc{5}{\it NP}\\
}}
\end{minipage}\\
\caption{
 CCG trees that are equally likely under Eq.~\ref{lew}.
 Our model resolves this ambiguity by modeling the head of every word (dependencies).
}
\label{example}
\end{figure}

Although much ambiguity is resolved with this supertagging, some ambiguity still remains.
Figure \ref{example} shows an example, where the two CCG parses are derived from the same supertags.
Lewis et al.'s approach to this problem is resorting to some deterministic rule.
For example, \newcite{lewis-lee-zettlemoyer:2016:N16-1} employ the attach low heuristics, which is motivated by the right-branching tendency of English, and always prioritizes (b) for this type of ambiguity.
Though for English it empirically works well, an obvious limitation is that it does not always derive the correct parse;
consider a phrase ``{\it a house in Paris with a garden}'', for which the correct parse has the structure corresponding to (a) instead.

In this paper, we provide a way to resolve these remaining ambiguities under the locally factored model, by explicitly modeling bilexical dependencies as shown in Figure \ref{example}.
Our joint model is still locally factored so that an efficient A* search can be applied.
The key idea is to predict the head of every word independently as in Eq.~\ref{lew} with a strong unigram model, for which we utilize the scoring model in the recent successful graph-based dependency parsing on LSTMs \cite{TACL885,DBLP:journals/corr/DozatM16}.
Specifically, we extend the bi-directional LSTM (bi-LSTM) architecture of \newcite{lewis-lee-zettlemoyer:2016:N16-1} predicting the supertag of a word
to predict the head of it at the same time with a bilinear transformation.

The importance of modeling structures beyond supertags is demonstrated in the performance gain in \newcite{lee-lewis-zettlemoyer:2016:EMNLP2016}, which adds a recursive component to the model of Eq.~\ref{lew}.
Unfortunately, this formulation loses the efficiency of the original one since it needs to compute a recursive neural network every time it searches for a new node.
Our model does not resort to the recursive networks while modeling tree structures via dependencies.

We also extend the tri-training method of \newcite{lewis-lee-zettlemoyer:2016:N16-1} to learn our model with dependencies from unlabeled data.
On English CCGbank test data,
our model with this technique achieves 88.8\% and 94.0\% in terms of labeled and unlabeled F1, which mark the best scores so far.

Besides English, we provide experiments on Japanese CCG parsing.
Japanese employs freer word order dominated by the case markers and a deterministic rule such as the attach low method may not work well.
We show that this is actually the case;
our method outperforms the simple application of \citet{lewis-lee-zettlemoyer:2016:N16-1} in a large margin, 10.0 points in terms of clause dependency accuracy.

\section{Background}

\label{related}
Our work is built on A* CCG parsing (Section \ref{astar}), which we extend in Section \ref{proposed} with a head prediction model on bi-LSTMs (Section \ref{bi-dep}).

\subsection{Supertag-factored A*~CCG Parsing}
\label{astar}
CCG has a nice property that since every category is highly informative about attachment decisions,
assigning it to every word~({\it supertagging}) resolves most of its syntactic structure.
\citet{lewis-steedman:2014:EMNLP2014} utilize this characteristics of the grammar.
Let a CCG tree ${\bm y}$ be a list of categories $\langle c_1, \ldots, c_N \rangle$ and a derivation on it.
Their model looks for the most probable ${\bm y}$ given a sentence ${\bm x}$ of length $N$ from the set $Y({\bm x})$ of possible CCG trees under the model of Eq.~\ref{lew}:
\begin{align}
 \hat{{\bm y}} = \argmax_{{\bm y} \in Y({\bm x})} \sum_{i \in [1,N]} \log P_{tag}(c_i | {\bm x}). \nonumber
\end{align}
Since this score is factored into each supertag, they call the model a {\it supertag-factored} model.

Exact inference of this problem is possible by A* parsing \cite{klein:a_star}, which uses the following two scores on a chart:
\begin{eqnarray*}
    b(C_{i,j}) & = & \sum_{c_k \in {\bm c}_{i,j}} \log P_{tag}(c_k | {\bm x}), \\
    a(C_{i,j}) & = & \sum_{k \in [1,N] \setminus [i, j]} \max_{c_k} \log P_{tag}(c_k | {\bm x}),
\end{eqnarray*}
where $C_{i,j}$ is a chart item called an {\it edge}, which abstracts parses spanning interval $[i,j]$ rooted by category $C$.
The chart maps each edge to the derivation with the highest score, i.e., the Viterbi parse for $C_{i,j}$.
${\bm c}_{i,j}$ is the sequence of categories on such Viterbi parse, and thus $b$ is called the Viterbi inside score, while $a$ is the approximation (upper bound) of the Viterbi outside score.

A* parsing is a kind of CKY chart parsing augmented with an agenda,
a priority queue that keeps the edges to be explored.
At every step it pops the edge $e$ with the highest priority $b(e) + a(e)$ and inserts that into the chart, and enqueue any edges that can be
built by combining $e$ with other edges in the chart.
The algorithm terminates when an edge $C_{1,N}$ is popped from the agenda.

A* search for this model is quite efficient because both $b$ and $a$ can be obtained from the unigram category distribution on every word, which can be precomputed before search.
The heuristics $a$ gives an upper bound on the true Viterbi outside score (i.e., admissible).
Along with this the condition that the inside score never increases by expansion (monotonicity) guarantees that the first found derivation on $C_{1,N}$ is always optimal.
$a(C_{i,j})$ matches the true outside score if the one-best category assignments on the outside words ($\arg\max_{c_k} \log P_{tag}(c_k | {\bm x})$) can comprise a well-formed tree with $C_{i,j}$, which is generally not true.

\paragraph{Scoring model}
For modeling $P_{tag}$, \citet{lewis-steedman:2014:EMNLP2014} use a log-linear model with features from a fixed window context.
\citet{lewis-lee-zettlemoyer:2016:N16-1} extend this with bi-LSTMs, which encode the complete sentence and capture the long range syntactic information.
We base our model on this bi-LSTM architecture, and extend it to modeling a head word at the same time.

\paragraph{Attachment ambiguity}
In A* search, an edge with the highest priority $b+a$ is searched first, but as shown in Figure \ref{example} the same categories (with the same priority) may sometimes derive more than one tree.
In \citet{lewis-steedman:2014:EMNLP2014}, they prioritize the parse with longer dependencies, which they judge with a conversion rule from a CCG tree to a dependency tree (Section \ref{conversion}).
\citet{lewis-lee-zettlemoyer:2016:N16-1} employ another heuristics prioritizing low attachments of constituencies, but inevitably these heuristics cannot be flawless in any situations.
We provide a simple solution to this problem by explicitly modeling bilexical dependencies.


\subsection{Bi-LSTM Dependency Parsing}
\label{bi-dep}
For modeling dependencies, we borrow the idea from the recent graph-based neural dependency parsing \cite{TACL885,DBLP:journals/corr/DozatM16} in which each dependency arc is scored directly on the outputs of bi-LSTMs.
Though the model is first-order, bi-LSTMs enable conditioning on the entire sentence and lead to the state-of-the-art performance.
Note that this mechanism is similar to modeling of the supertag distribution discussed above, in that for each word the distribution of the head choice is unigram and can be precomputed.
As we will see this keeps our joint model still locally factored and A* search tractable.
For score calculation, we use an extended bilinear transformation by \newcite{DBLP:journals/corr/DozatM16} that models the prior headness of each token as well, which they call {\it biaffine}.



\section{Proposed Method}
\label{proposed}
\subsection{A* parsing with Supertag and Dependency Factored Model}
\label{proposedastar}

We define a CCG tree ${\bm y}$ for a sentence ${\bm x}=\langle x_i, \ldots, x_N \rangle$
as a triplet of a list of CCG categories ${\bm c}=\langle c_1, \ldots, c_N \rangle$,
dependencies ${\bm h}=\langle h_1, \ldots, h_N \rangle$, and the derivation,
where $h_i$ is the head index of $x_i$.
Our model is defined as follows:
\begin{align}
    \label{our_model}
    P({\bm y} | {\bm x}) & = \prod_{i \in [1,N]} P_{tag}(c_i | {\bm x}) \prod_{i \in [1,N]} P_{dep}(h_i | {\bm x}).
\end{align}
The added term $P_{dep}$ is a unigram distribution of the head choice.

A* search is still tractable under this model.
The search problem is changed as:
\begin{align}
    & \hat{{\bm y}} = \argmax_{{\bm y} \in Y({\bm x})} \Biggl( \sum_{i \in [1,N]} \log P_{tag}(c_i | {\bm x})  \nonumber\\
     &~~~~~~~~~~~~~~~~~~~~+  \sum_{i \in [1,N]} \log P_{dep}(h_i | {\bm x}) \Biggr),\nonumber
\end{align}
and the inside score is given by:
\begin{align}
    b(C_{i,j}) & = \sum_{c_k \in {\bm c}_{i,j}} \log P_{tag}(c_k | {\bm x}) \label{eq:newinside} \\
    &+ \sum_{k \in [i,j]\setminus\{root({\bm h}^C_{i,j})\}} \log P_{dep}(h_k | {\bm x}), \nonumber
\end{align}
where ${\bm h}^C_{i,j}$ is a dependency subtree for the Viterbi parse on $C_{i,j}$ and $root({\bm h})$ returns the root index.
We exclude the head score for the subtree root token since it cannot be resolved inside $[i,j]$.
This causes the mismatch between the goal inside score $b(C_{1,N})$ and the true model score (log of Eq.~\ref{our_model}), which we adjust by adding a special unary rule that is always applied to the popped goal edge $C_{1,N}$.

\begin{figure}[t]
\centering
\includegraphics[bb=0 0 479 195,width=7cm]{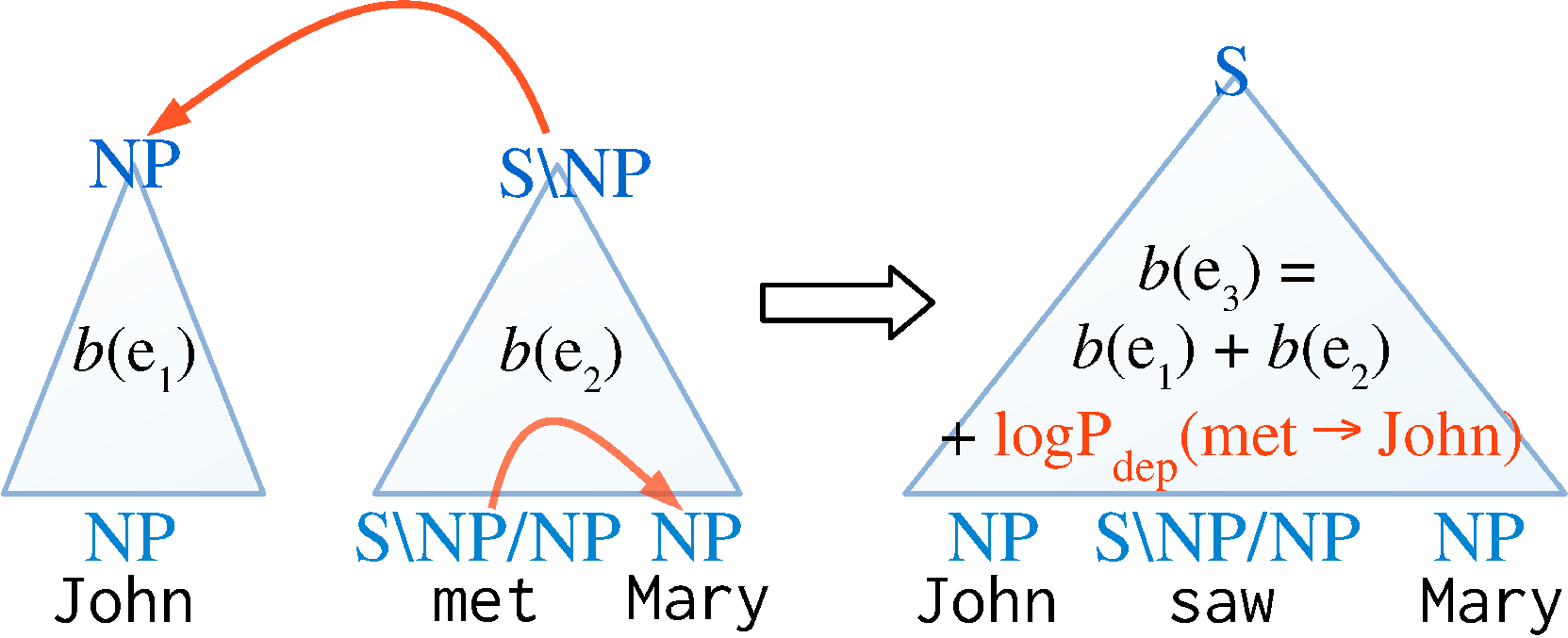}
    \caption{Viterbi inside score for edge $e_3$ under our model is
    the sum of those of $e_1$ and $e_2$ and the score of dependency arc going
    from the head of $e_2$ to that of $e_1$ (the head direction changes according to the child categories).}
\label{calc}
\end{figure}

We can calculate the dependency terms in Eq.~\ref{eq:newinside} on the fly when expanding the chart.
Let the currently popped edge be $A_{i,k}$, which will be combined with $B_{k,j}$ into $C_{i,j}$.
The key observation is that only one dependency arc (between $root({\bm h}_{i,k}^A)$ and $root({\bm h}_{k,j}^B)$) is resolved at every combination (see Figure \ref{calc}).
For every rule $C \rightarrow A~B$ we can define the head direction (see Section \ref{conversion}) and $P_{dep}$ is obtained accordingly.
For example, when the right child $B$ becomes the head,
$b(C_{i,j}) = b(A_{i,k}) + b(B_{k,j}) + \log P_{dep}(h_{l}=m|{\bm x})$, where $l=root({\bm h}_{i,k}^A)$ and $m=root({\bm h}_{k,j}^B)$ ($l < m$).

The Viterbi outside score is changed as:
\begin{align}
    &a(C_{i,j}) = \sum_{k \in [1,N] \setminus [i, j]} \max_{c_k} \log P_{tag}(c_k | {\bm x}) \nonumber \\
    &~~~~~~~~~+~~~~~~~\sum_{k \in L} \max_{h_k} \log P_{dep}(h_k | {\bm x}), \nonumber
\end{align}
where $L = [1,N] \setminus [k' | k' \in [i,j], root({\bm h}_{i,j}^C) \neq k']$.
We regard $root({\bm h}_{i,j}^C)$ as an outside word since its head is undefined yet.
For every outside word we independently assign the weight of its argmax head, which may not comprise a well-formed dependency tree.
We initialize the agenda by adding an item for every supertag $C$ and word $x_i$ with the score
    $a(C_{i,i}) = \sum_{k \in I\setminus \{i\}} \max \log P_{tag}(c_k | {\bm x})
     + \sum_{k \in I} \max \log P_{dep}(h_k | {\bm x})$.
Note that the dependency component of it is the same for every word.

\subsection{Network Architecture}
\begin{figure}[t]
\centering
\includegraphics[bb=0 0 445 397,width=6cm]{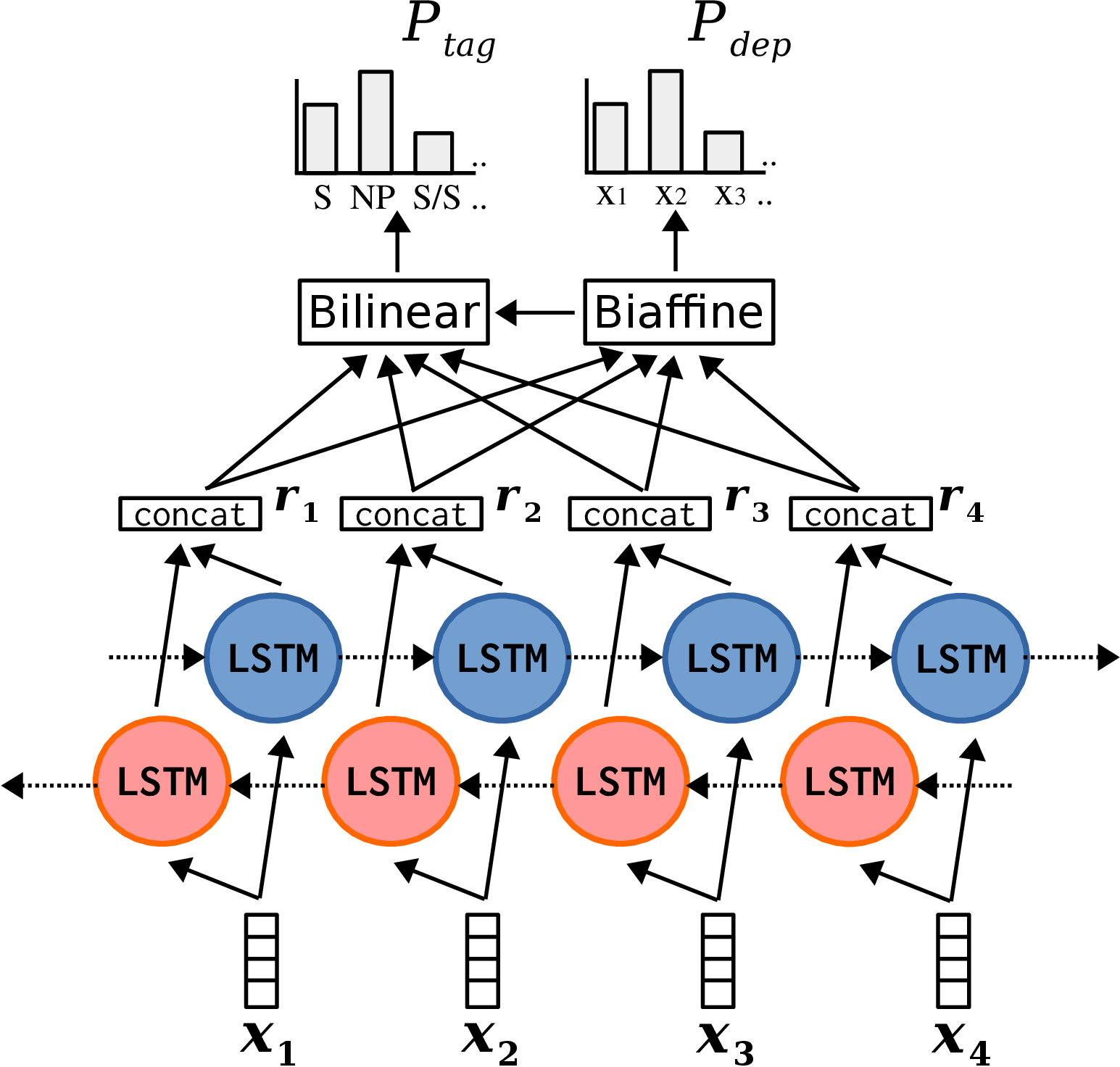}
    \caption{Neural networks of our supertag and
    dependency factored model. First we map every word $x_i$ to
    a hidden vector ${\bm r}_i$ by bi-LSTMs,
    and then apply biaffine~(Eq.~\ref{biaffine}) and bilinear~(Eq.~\ref{bilinear}) transformations
    to obtain the distributions of dependency heads ($P_{dep}$) and supertags ($P_{tag}$).}
\label{arch}
\end{figure}

Following \citet{lewis-lee-zettlemoyer:2016:N16-1} and \citet{DBLP:journals/corr/DozatM16},
we model $P_{tag}$ and $P_{dep}$ using bi-LSTMs for exploiting the entire sentence to capture the long range phenomena.
See Figure \ref{arch} for the overall network architecture, where $P_{tag}$ and $P_{dep}$ share the common bi-LSTM hidden vectors.

First we map every word $x_i$ to their hidden vector ${\bm r}_i$ with bi-LSTMs.
The input to the LSTMs is word embeddings, which we describe in Section~\ref{experiment}.
We add special start and end tokens to each sentence with the trainable
parameters following \citet{lewis-lee-zettlemoyer:2016:N16-1}.
For $P_{dep}$, we use the biaffine transformation in \citet{DBLP:journals/corr/DozatM16}:
\begin{align}
    &{\bm g}_i^{dep} = MLP_{child}^{dep}({\bm r}_i), \nonumber \\
    &{\bm g}_{h_i}^{dep} = MLP_{head}^{dep}({\bm r}_{h_i}), \nonumber \\
    &P_{dep}(h_i | {\bm x}) \label{biaffine} \\
    &~~~~~~~~ \propto \exp(({\bm g}_i^{dep})^{\mathrm{T}} W_{dep} {\bm g}_{h_i}^{dep}
    + {\bm w}_{dep} {\bm g}_{h_i}^{dep}), \nonumber
\end{align}
where $MLP$ is a multilayered perceptron.
Though \citet{lewis-lee-zettlemoyer:2016:N16-1} simply use an
MLP for mapping ${\bm r}_i$ to $P_{tag}$,
we additionally utilize the hidden vector of
the most probable head
$h_i = \arg\max_{h_i'} P_{dep}(h_i' | {\bm x})$,
and apply ${\bm r}_i$
and ${\bm r}_{h_i}$ to a bilinear function:\footnote{
This is inspired by the formulation of label prediction
in \citet{DBLP:journals/corr/DozatM16}, which
performs the best among other settings that remove or reverse the
dependence between the head model and the supertag model.
}
\begin{align}
    &{\bm g}_i^{tag} = MLP_{child}^{tag}({\bm r}_i), \nonumber \\
    &{\bm g}_{h_i}^{tag} = MLP_{head}^{tag}({\bm r}_{h_i}), \label{bilinear} \\
    &{\bm \ell} = ({\bm g}_i^{tag})^{\mathrm{T}} {\bm U}_{tag} {\bm g}_{h_i}^{tag} +
     W_{tag}
        \begin{bmatrix}
            {\bm g}_i^{tag} \\ {\bm g}_{h_i}^{tag}
        \end{bmatrix} + {\bm b}_{tag}, \nonumber \\
    &P_{tag}(c_i | {\bm x}) \propto \exp(\ell_c), \nonumber
\end{align}
where ${\bm U}_{tag}$ is a third order tensor.
As in Lewis et al. these values can be precomputed before search, which makes our A* parsing quite efficient.

\begin{figure*}[t]
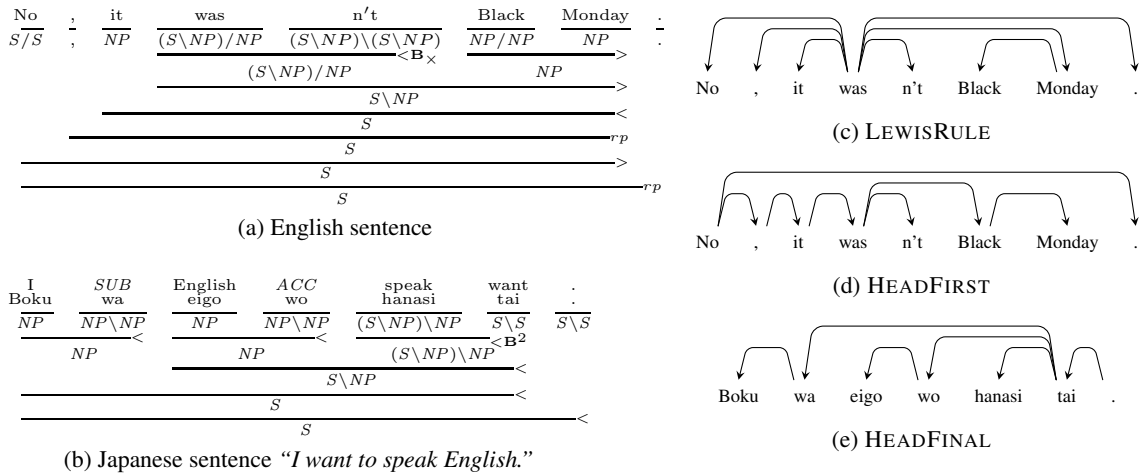

\begin{minipage}[t]{0.5\hsize}
    \tiny
    \subfloat[English sentence \label{englishsentence}]{
\deriv{8}{
{\rm No}&{\rm ,}& {\rm it}& {\rm was}& {\rm n't}& {\rm Black}& {\rm Monday}& {\rm .}\\
\uline{1}& \uline{1}    & \uline{1} & \uline{1}& \uline{1} & \uline{1} & \uline{1}& \uline{1}\\
\it S/S &\it , &\it  NP &\it (S\bs NP)/NP & (S\bs NP)\bs (S\bs NP) &\it NP/NP &\it NP &\it .\\
    &&&\bxcomp{2} & \fapply{2}\\
&&&\mcc{2}{\it (S\bs NP)/NP} & \mcc{2}{\it NP}\\
    &&&\fapply{4}\\
&&&\mcc{4}{\it S\bs NP}\\
    &&\bapply{5}\\
&&\mcc{5}{\it S}\\
&\comb{6}{rp}\\
&\mcc{6}{\it S}\\
    \fapply{7}\\
\mcc{7}{\it S}\\
\comb{8}{rp}\\
\mcc{8}{\it S}\\
}}\\[0.5ex]

    \subfloat[Japanese sentence {\it ``I want to speak English.''}  \label{sub:iwanttospeak}]{
\deriv{7}{
{\rm I}&{\it SUB}& {\rm English}& {\it ACC}& {\rm speak}& {\rm want}& {\rm .}\\
{\rm Boku}&{\rm wa}& {\rm eigo}& {\rm wo}& {\rm hanasi}& {\rm tai}& {\rm .}\\
\uline{1}& \uline{1}& \uline{1} & \uline{1}& \uline{1} & \uline{1} & \uline{1}\\
\it NP &\it NP\bs NP &\it NP &\it NP\bs NP&\it (S\bs NP)\bs NP &\it S\bs S &\it S\bs S\\
    \bapply{2} & \bapply{2} & \bcomptwo{2} \\
    \mcc{2}{\it NP} &\mcc{2}{\it NP}&\mcc{2}{\it (S\bs NP)\bs NP}\\
    &&\bapply{4}\\
    &&\mcc{4}{\it S\bs NP}\\
\bapply{6}\\
\mcc{6}{\it S}\\
\bapply{7}\\
\mcc{7}{\it S}\\
}}
\end{minipage}
    \hfill
\begin{minipage}[t]{0.5\hsize}
    \small
\centering
    \subfloat[{\sc LewisRule} \label{lewisrule}]{
\begin{dependency}
   \begin{deptext}[column sep=1.em, font=\scriptsize]
      No \& , \& it \& was \& n't \& Black \& Monday \& . \\
   \end{deptext}
   \depedge[hide label, edge height=3ex, edge horizontal padding=0.5ex]{4}{5}{}
   \depedge[hide label, edge height=4ex, edge horizontal padding=0.5ex]{4}{7}{}
   \depedge[hide label, edge height=5ex, edge horizontal padding=0.5ex]{4}{8}{}
   \depedge[hide label, edge height=3ex]{7}{6}{}
   \depedge[hide label, edge height=3ex, edge horizontal padding=0.5ex]{4}{3}{}
   \depedge[hide label, edge height=4ex, edge horizontal padding=0.5ex]{4}{2}{}
   \depedge[hide label, edge height=5ex, edge horizontal padding=0.5ex]{4}{1}{}
\end{dependency}}\\[1ex]

    \subfloat[{\sc HeadFirst} \label{headfirst}]{
\begin{dependency}
   \begin{deptext}[column sep=1.em, font=\scriptsize]
      No \& , \& it \& was \& n't \& Black \& Monday \& . \\
   \end{deptext}
   \depedge[hide label, edge height=3ex]{4}{5}{}
   \depedge[hide label, edge height=4ex]{4}{6}{}
   \depedge[hide label, edge height=5ex]{1}{8}{}
   \depedge[hide label, edge height=3ex]{6}{7}{}
   \depedge[hide label, edge height=3ex]{3}{4}{}
   \depedge[hide label, edge height=3ex]{2}{3}{}
   \depedge[hide label, edge height=3ex]{1}{2}{}
\end{dependency}}\\[1ex]

    \subfloat[{\sc HeadFinal} \label{sub:headfinal}]{
\begin{dependency}
   \begin{deptext}[column sep=1.em, font=\scriptsize]
      Boku \& wa \& eigo \& wo \& hanasi \& tai \& . \\
   \end{deptext}
   \depedge[hide label, edge height=3ex]{2}{1}{}
   \depedge[hide label, edge height=3ex]{4}{3}{}
   \depedge[hide label, edge height=3ex]{6}{5}{}
   \depedge[hide label, edge height=4ex]{6}{4}{}
   \depedge[hide label, edge height=5ex]{6}{2}{}
   \depedge[hide label, edge height=3ex]{7}{6}{}
\end{dependency}}
\end{minipage}
\caption{Examples of applying conversion rules in Section~\ref{conversion} to English and Japanese sentences.}
\label{no_it_was}
\end{figure*}

\section{CCG to Dependency Conversion}
\label{conversion}
Now we describe our conversion rules from a CCG tree to a dependency one, which we use in two purposes:
1) creation of the training data for the dependency component of our model;
and 2) extraction of a dependency arc at each combinatory rule during A* search (Section \ref{proposedastar}).
\citet{lewis-steedman:2014:EMNLP2014} describe one way to extract dependencies from a CCG tree ({\sc LewisRule}).
Below in addition to this we describe two simpler alternatives ({\sc HeadFirst} and {\sc HeadFinal}),
and see the effects on parsing performance in our experiments (Section \ref{experiment}).
See Figure~\ref{no_it_was} for the overview.

\paragraph{\bfseries{\scshape{LewisRule}}}
This is the same as the conversion rule in \citet{lewis-steedman:2014:EMNLP2014}.
As shown in Figure \ref{lewisrule} the output looks a familiar English dependency tree.

For forward application and (generalized) forward composition,
we define the head to be the left argument of the combinatory rule,
unless it matches either $X/X$ or $X/(X\backslash Y)$,
in which case the right argument is the head.
For example, on {\it ``Black Monday''} in Figure~\ref{englishsentence} we choose {\it Monday} as the head of {\it Black}.
For the backward rules, the conversions are defined as the reverse of
the corresponding forward rules.
For other rules, {\it RemovePunctuation}~({\it rp}) chooses the non punctuation argument as the head, while {\it Conjunction}~($\Phi$) chooses the right argument.\footnote{When
applying {\sc LewisRule} to Japanese,
we ignore the feature values in determining the head argument,
which we find often leads to a more natural dependency structure.
For example, in ``tabe ta'' (eat {\it PAST}), the category of
auxiliary verb ``ta'' is $S_{f_1}\backslash S_{f_2}$ with $f_1 \neq f_2$,
and thus $S_{f_1} \neq S_{f_2}$.
We choose ``tabe'' as the head in this case by removing the feature values,
which makes the category $X\backslash X$.
}

One issue when applying this method for obtaining the training data is that due to the mismatch between the rule set of our CCG parser, for which we follow \citet{lewis-steedman:2014:EMNLP2014}, and the grammar in English CCGbank \cite{ccgbank} we cannot extract dependencies from some of annotated CCG trees.\footnote{
For example, the combinatory rules in \citet{lewis-steedman:2014:EMNLP2014} do not contain $N_{conj} \to N~N$ in CCGbank.
Another difficulty is that in English CCGbank the name of each combinatory rule is not annotated explicitly.}
For this reason, we instead obtain the training data for this method from the original dependency annotations on CCGbank.
Fortunately the dependency annotations of CCGbank matches {\sc LewisRule} above in most cases and thus they can be a good approximation to it.

\paragraph{\bfseries{\scshape{HeadFinal}}}
Among SOV languages, Japanese is known as a strictly head final language, meaning that the head of every word always follows it.
Japanese dependency parsing \cite{E99-1026,DBLP:conf/conll/KudoM02} has exploited this property explicitly by only allowing left-to-right dependency arcs.
Inspired by this tradition, we try a simple {\sc HeadFinal} rule in Japanese CCG parsing, in which we always select the right argument as the head.
For example we obtain the head final dependency tree in Figure \ref{sub:headfinal} from the Japanese CCG tree in Figure \ref{sub:iwanttospeak}.

\paragraph{\bfseries{\scshape{HeadFirst}}}
We apply the similar idea as {\sc HeadFinal} into English.
Since English has the opposite, SVO word order, we define the simple ``head first'' rule, in which the left argument always becomes the head (Figure~\ref{headfirst}).

Though this conversion may look odd at first sight it also has some advantages over {\sc LewisRule}.
First, since the model with {\sc LewisRule} is trained on the CCGbank dependencies, at inference, occasionally the two components $P_{dep}$ and $P_{tag}$ cause some conflicts on their predictions.
For example, the true Viterbi parse may have a lower score in terms of dependencies, in which case the parser slows down and may degrade the accuracy.
{\sc HeadFirst}, in contract, does not suffer from such conflicts.
Second, by fixing the direction of arcs, the prediction of heads becomes easier, meaning that the dependency predictions become more reliable.
Later we show that this is in fact the case for existing dependency parsers (see Section \ref{sec:tritrain}), and in practice, we find that this simple conversion rule leads to the higher parsing scores than {\sc LewisRule} on English (Section \ref{experiment}).

\section{Tri-training}
\label{sec:tritrain}
We extend the existing tri-training method to our models and apply it to our English parsers.

Tri-training
is one of the semi-supervised methods,
in which the outputs of two parsers on unlabeled data
are intersected to create (silver) new training data.
This method is successfully applied to dependency parsing~\cite{weiss-EtAl:2015:ACL-IJCNLP}
and CCG supertagging~\cite{lewis-lee-zettlemoyer:2016:N16-1}.

We simply combine the two previous approaches.
\citet{lewis-lee-zettlemoyer:2016:N16-1} obtain their silver data annotated with the high quality supertags.
Since they make this data publicly available \footnote{\url{https://github.com/uwnlp/taggerflow}}, we obtain our silver data by
assigning dependency structures on top of them.\footnote{We annotate POS tags on this data using Stanford POS tagger \cite{N03-1033}.}

We train two very different dependency parsers from the training data extracted from CCGbank Section 02-21.
This training data differs depending on our dependency conversion strategies (Section \ref{conversion}).
For {\sc LewisRule}, we extract the original dependency annotations of CCGbank.
For {\sc HeadFirst}, we extract the head first dependencies from the CCG trees.
Note that we cannot annotate dependency labels so we assign a dummy ``none'' label to every arc.
The first parser is graph-based {\tt RBGParser}~\cite{P14-1130}
with the default settings except that we train an unlabeled parser
and use word embeddings of \citet{P10-1040}.
The second parser is
transition-based {\tt lstm-parser}~\cite{P15-1033} with the default parameters.

On the development set (Section 00),
with {\sc LewisRule} dependencies
{\tt RBGParser} shows 93.8\% unlabeled attachment score while that of {\tt lstm-parser} is 92.5\% using gold POS tags.
Interestingly, the parsers with {\sc HeadFirst} dependencies achieve higher scores:
94.9\% by {\tt RBGParser} and 94.6\% by {\tt lstm-parser}, suggesting that {\sc HeadFirst} dependencies are easier to parse.
For both dependencies, we obtain more than 1.7 million
sentences on which two parsers agree.

Following \citet{lewis-lee-zettlemoyer:2016:N16-1},
we include 15 copies of CCGbank training set when using these silver data.
Also to make effects of the tri-train samples smaller we multiply their loss by 0.4.

\section{Experiments}

\newcommand*\cpp{C\kern-0.2ex\raisebox{0.4ex}{\scalebox{0.8}{+\kern-0.4ex+}}}
\newcommand*{\brokenurl}[2]{\href{#1#2}{\texttt{#1}}\par\nopagebreak\href{#1#2}{\texttt{#2}}}
\label{experiment}
We perform experiments on English and Japanese CCGbanks.

\subsection{English Experimental Settings}
We follow the standard data splits and use Sections 02-21 for training, Section 00 for development, and Section 23 for final evaluation.
We report labeled and unlabeled F1 of the extracted CCG semantic dependencies obtained using {\tt generate} program supplied with {\tt C\&C} parser.

For our models, we adopt the pruning strategies in \citet{lewis-steedman:2014:EMNLP2014} and
allow at most 50 categories per word,
use a variable-width beam with $\beta=0.00001$,
and utilize a tag dictionary, which maps frequent words to the possible supertags\footnote{We use
the same tag dictionary provided with their bi-LSTM model.}.
Unless otherwise stated, we only allow normal form parses~\cite{P96-1011,C10-1053}, choosing the same subset of the constraints as \citet{lewis-steedman:2014:EMNLP2014}.

We use as word representation the concatenation of word vectors
initialized to GloVe\footnote{\url{http://nlp.stanford.edu/projects/glove/}}~\cite{pennington2014glove},
and randomly initialized prefix and suffix vectors of the length 1 to 4,
which is inspired by \citet{lewis-lee-zettlemoyer:2016:N16-1}.
All affixes appearing less than two times in the training data are mapped to ``UNK''.

Other model configurations are:
4-layer bi-LSTMs with
left and right 300-dimensional LSTMs,
1-layer 100-dimensional MLPs with ELU non-linearity~\cite{DBLP:journals/corr/ClevertUH15}
for all $MLP_{child}^{dep}$,~$MLP_{head}^{dep}$,~$MLP_{child}^{tag}$ and $MLP_{head}^{tag}$,
and the Adam optimizer with $\beta_1=0.9, \beta_2=0.9$, L2 norm ($1e^{-6}$), and learning rate decay with the ratio 0.75 for every 2,500 iteration
starting from $2e^{-3}$,
which is shown to be effective for training the biaffine parser~\cite{DBLP:journals/corr/DozatM16}.

\subsection{Japanese Experimental Settings}
We follow the default train/dev/test splits of Japanese CCGbank \cite{uematsu-EtAl:2013:ACL2013}.
For the baselines, we use an existing shift-reduce CCG parser implemented in an NLP tool Jigg\footnote{\url{https://github.com/mynlp/jigg}} \cite{noji-miyao:2016:P16-4}, and our implementation of the supertag-factored model using bi-LSTMs.

For Japanese, we use as word representation
the concatenation of word vectors
initialized to Japanese Wikipedia Entity Vector\footnote{
    \brokenurl{http://www.cl.ecei.tohoku.ac.jp/}{~m-suzuki/jawiki\_vector/}},
and 100-dimensional vectors computed from randomly initialized
50-dimensional character embeddings through convolution~\cite{santos:ICML}.
We do not use affix vectors as affixes are less informative in Japanese.
All characters appearing less than two times are mapped to ``UNK''.
We use the same parameter settings as English for bi-LSTMs, MLPs, and optimization.

One issue in Japanese experiments is evaluation.
The Japanese CCGbank is encoded in a different format than the English bank, and no standalone script for extracting semantic dependencies is available yet.
For this reason, we evaluate the parser outputs by converting them to {\it bunsetsu dependencies}, the syntactic representation ordinary used in Japanese NLP \cite{DBLP:conf/conll/KudoM02}.
Given a CCG tree, we obtain this by first segment a sentence into bunsetsu (chunks) using CaboCha\footnote{\url{http://taku910.github.io/cabocha/}} and extract dependencies that cross a bunsetsu boundary after obtaining the word-level, head final dependencies as in Figure \ref{sub:iwanttospeak}.
For example, the sentence in Figure \ref{sub:headfinal} is segmented as ``{\it Boku wa $|$ eigo wo $|$ hanashi tai}'', from which we extract two dependencies ({\it Boku wa}) $\leftarrow$ ({\it hanashi tai}) and ({\it eigo wo}) $\leftarrow$ ({\it hanashi tai}).
We perform this conversion for both gold and output CCG trees and calculate the (unlabeled) attachment accuracy.
Though this is imperfect, it can detect important parse errors such as attachment errors and thus can be a good proxy for the performance as a CCG parser.

\begin{table}[t]
    \centering
\scalebox{0.9}{
\begin{tabular}{lcc} \hline
    \multicolumn{1}{c}{{\bf Method}} & {\bf Labeled} & {\bf Unlabeled} \\ \hline
    {\it CCGbank}   \\
    ~~{\sc LewisRule}  w/o dep &  85.8 & 91.7 \\
    ~~{\sc LewisRule}   &  86.0 & 92.5 \\
    ~~{\sc HeadFirst}  w/o dep &  85.6 & 91.6 \\
    ~~{\sc HeadFirst}   &  {\bf 86.6} & {\bf 92.8} \\ \hline \hline
    {\it Tri-training}   \\
    ~~{\sc LewisRule}  & 86.9  & 93.0 \\
    ~~{\sc HeadFirst}  & {\bf 87.6} & {\bf 93.3} \\ \hline
\end{tabular}}
    \caption{Parsing results (F1) on English development set.
 ``w/o dep'' means that the model discards dependency components at prediction.
 }
\label{english_experiment_dev}
\end{table}

\begin{table}[t]
    \centering
\scalebox{0.78}{
\begin{tabular}{lccc} \hline
    \multicolumn{1}{c}{{\bf Method}} & {\bf Labeled} & {\bf Unlabeled} & {\bf \# violations} \\ \hline
    {\it CCGbank}   \\
    ~~{\sc LewisRule} w/o dep &  85.8 & 91.7 & 2732 \\
    ~~{\sc LewisRule}  &  85.4 & 92.2 & 283 \\
    ~~{\sc HeadFirst} w/o dep &  85.6 & 91.6 & 2773 \\
    ~~{\sc HeadFirst}  & {\bf 86.8} &  {\bf 93.0}  & {\bf 89} \\ \hline \hline
    {\it Tri-training}   \\
    ~~{\sc LewisRule}  & 86.7  & 92.8 & 253 \\
    ~~{\sc HeadFirst}  & {\bf 87.7} & {\bf 93.5} & {\bf 66} \\ \hline
\end{tabular}}
    \caption{Parsing results (F1) on English development set when excluding the normal form constraints.
 \# violations is the number of combinations violating the constraints on the outputs.}
\label{english_experiment_dev_wo_normalform}
\end{table}

%

\begin{table}[t]
    \centering
\scalebox{0.78}{
\begin{tabular}{lcc} \hline
    \multicolumn{1}{c}{{\bf Method}} & {\bf Labeled} & {\bf Unlabeled} \\ \hline
    {\it CCGbank}   \\
    ~~{\tt C\&C}~\cite{J07-4004}  &  85.5 & 91.7  \\
    ~~~~w/ LSTMs~\cite{N16-1027}  &  {\bf 88.3} & - \\
    ~~{\tt EasySRL}~\cite{lewis-lee-zettlemoyer:2016:N16-1} &  87.2 & -  \\
    ~~{\tt EasySRL\_reimpl} &  86.8 & 92.3  \\
    ~~{\sc HeadFirst} w/o NF~(Ours)  & 87.7 & {\bf 93.4} \\ \hline \hline
    {\it Tri-training}   \\
    ~~{\tt EasySRL}~\cite{lewis-lee-zettlemoyer:2016:N16-1} &  88.0 & 92.9 \\
    ~~{\tt neuralccg}~\cite{lee-lewis-zettlemoyer:2016:EMNLP2016}  &  88.7 & 93.7 \\
    ~~{\sc HeadFirst} w/o NF~(Ours) &  {\bf 88.8} & {\bf 94.0} \\ \hline
\end{tabular}}
    \caption{Parsing results (F1) on English test set~(Section 23).}
\label{english_experiment}
\end{table}

\subsection{English Parsing Results}
\paragraph{Effect of Dependency}
We first see how the dependency components added in our model affect the performance.
Table~\ref{english_experiment_dev} shows the results on the development set with the several configurations, in which
``w/o dep'' means discarding the dependency terms of the model and applying the attach low heuristics (Section \ref{intro}) instead (i.e., a supertag-factored model; Section \ref{astar}).
We can see that for both {\sc LewisRule} and {\sc HeadFirst}, adding dependency terms improves the performance.


\paragraph{Choice of Dependency Conversion Rule}
To our surprise, our simple {\sc HeadFirst} strategy always leads to better results than the linguistically motivated {\sc LewisRule}.
The absolute improvements by tri-training are equally large (about 1.0 points), suggesting that our model with dependencies can also benefit from the silver data.


\paragraph{Excluding Normal Form Constraints}
One advantage of {\sc HeadFirst} is that the direction of arcs is always right, making the structures simpler and more parsable (Section \ref{sec:tritrain}).
From another viewpoint, this fixed direction means that the constituent structure behind a (head first) dependency tree is unique.
Since the constituent structures of CCGbank trees basically follow the normal form (NF), we hypothesize that the model learned with {\sc HeadFirst} has an ability to force the outputs in NF automatically.
We summarize the results without the NF constraints in Table \ref{english_experiment_dev_wo_normalform}, which shows that the above argument is correct;
the number of violating NF rules on the outputs of {\sc HeadFirst} is much smaller than that of {\sc LewisRule} (89 vs.~283).
Interestingly the scores of {\sc HeadFirst} slightly increase from the models with NF (e.g., 86.8 vs.~86.6 for CCGbank), suggesting that the NF constraints hinder the search of {\sc HeadFirst} models occasionally.


\paragraph{Results on Test Set}
Parsing results on the test set~(Section 23) are shown
in Table~\ref{english_experiment}, where we compare
our best performing {\sc HeadFirst} dependency model
without NF constraints with the several existing parsers.
In the CCGbank experiment, our parser shows the better result than
all the baseline parsers except C\&C with an LSTM supertagger~\cite{N16-1027}.
Our parser outperforms {\tt EasySRL} by 0.5\% and our reimplementation of that parser ({\tt EasySRL\_reimpl}) by 0.9\% in terms of labeled F1.
In the tri-training experiment, our parser shows much increased performance
of 88.8\% labeled F1 and 94.0\% unlabeled F1,
outperforming the current state-of-the-art {\tt neuralccg} \cite{lee-lewis-zettlemoyer:2016:EMNLP2016} that uses recursive neural networks
by 0.1 point and 0.3 point in terms of labeled and unlabeled F1.
This is the best reported F1 in English CCG parsing.
\begin{table}[t]
    \small
    \centering
    \begin{tabular}{lccc} \hline
        & {\tt EasySRL\_reimpl} & {\tt neuralccg} & Ours \\ \hline
        {\it Tagging} & 24.8  & 21.7 & 16.6 \\
        {\it A* Search} & 185.2 & 16.7 & 114.6 \\
        {\it Total} & 21.9 & 9.33 & 14.5 \\ \hline
\end{tabular}
    \caption{Results of the efficiency experiment, where each number is the number of sentences processed per second. We compare our proposed parser against {\tt neuralccg}
    and our reimplementation of {\tt EasySRL}.}
\label{efficiency_experiment}
\end{table}

\paragraph{Efficiency Comparison}
We compare the efficiency of our parser
with {\tt neuralccg} and {\tt EasySRL\_reimpl}.\footnote{This
experiment is performed on a laptop with 4-thread 2.0 GHz CPU.}
The results are shown in Table~\ref{efficiency_experiment}.
For the overall speed~(the third row), our parser is faster than {\tt neuralccg} although lags behind {\tt EasySRL\_reimpl}.
Inspecting the details, our supertagger runs slower than those of {\tt neuralccg} and {\tt EasySRL\_reimpl},
while in A* search our parser processes over 7 times more sentences than {\tt neuralccg}.
The delay in supertagging can be attributed to several factors, in particular
the differences in network architectures including the number of
bi-LSTM layers~(4 vs.~2) and the use of bilinear transformation
instead of linear one.
There are also many implementation differences in our parser~
(C\texttt{++} A* parser with neural network model implemented with Chainer~\cite{chainer_learningsys2015})
and {\tt neuralccg}~(Java parser with C\texttt{++} TensorFlow~\cite{tensorflow2015-whitepaper}
supertagger and
recursive neural model in C\texttt{++} DyNet~\cite{dynet}).


\subsection{Japanese Parsing Result}
We show the results of the Japanese parsing experiment in
Table~\ref{japanese_experiment}.
The simple application of \citet{lewis-lee-zettlemoyer:2016:N16-1} (Supertag model) is not effective for Japanese, showing the lowest attachment score of 81.5\%.
We observe a performance boost with our method, especially with {\sc HeadFinal} dependencies, which outperforms the baseline shift-reduce parser
by 1.1 points on category assignments and 4.0 points on bunsetsu dependencies.

\begin{figure}[t]
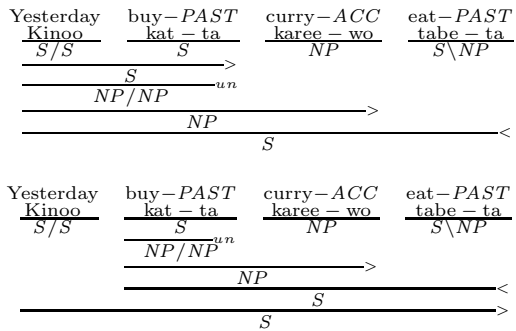

\centering
\fontsize{7.0pt}{0pt}\selectfont
\deriv{4}{
{\rm Yesterday}&{\rm buy-}{\it PAST}& {\rm curry-}{\it ACC}& {\rm eat-}{\it PAST}\\
{\rm Kinoo}&{\rm kat-ta}& {\rm karee-wo}& {\rm tabe-ta}\\
\uline{1}& \uline{1}    & \uline{1} & \uline{1}\\
\it S/S   &\it S &\it  NP &\it S\bs NP \\
\fapply{2}\\
\mcc{2}{\it S}\\
\comb{2}{un}\\
\mcc{2}{\it NP/NP}\\
\fapply{3}\\
\mcc{3}{\it NP}\\
\bapply{4}\\
\mcc{4}{\it S}\\
}\\[4ex]
\deriv{4}{
{\rm Yesterday}&{\rm buy-}{\it PAST}& {\rm curry-}{\it ACC}& {\rm eat-}{\it PAST}\\
{\rm Kinoo}&{\rm kat-ta}& {\rm karee-wo}& {\rm tabe-ta}\\
\uline{1}& \uline{1}    & \uline{1} & \uline{1}\\
\it S/S   &\it S &\it  NP &\it S\bs NP \\
&\comb{1}{un}\\
&\mcc{1}{\it NP/NP}\\
&\fapply{2}\\
&\mcc{2}{\it NP}\\
&\bapply{3}\\
&\mcc{3}{\it S}\\
\fapply{4}\\
\mcc{4}{\it S}\\
}
\caption{Examples of ambiguous Japanese sentence given fixed supertags.
 The English translation is {\it ``I ate the curry I bought yesterday''}.}
\label{ja_example}
\end{figure}

The degraded results of the simple application of the supertag-factored
model can be attributed to
the fact that the structure of a Japanese sentence is still highly ambiguous
given the supertags~(Figure~\ref{ja_example}).
This is particularly the case in constructions
where phrasal adverbial/adnominal modifiers (with the supertag $S/S$) are involved.
The result suggests the importance of modeling dependencies in some languages, at least Japanese.

\begin{table}[t]
    \small
    \centering
\begin{tabular}{lcc} \hline
    \multicolumn{1}{c}{{\bf Method}} & {\bf Category} & {\bf Bunsetsu Dep.} \\ \hline
    \citet{noji-miyao:2016:P16-4} & 93.0 & 87.5 \\
    Supertag model & 93.7 & 81.5 \\
    {\sc LewisRule}~(Ours) & 93.8 & 90.8 \\
    {\sc HeadFinal}~(Ours) & {\bf 94.1} & {\bf 91.5} \\ \hline
\end{tabular}
\caption{Results of Japanese CCGbank.}
\label{japanese_experiment}
\end{table}

\section{Related Work}
\label{sec:related}

There is some past work that utilizes dependencies in lexicalized grammar parsing,
which we review briefly here.

For Head-driven Phrase Structure Grammar~(HPSG; \citet{pollard1994head}),
there are studies to use the predicted dependency structure
to improve HPSG parsing accuracy.
\citet{P07-1079} use dependencies to constrain the form of the output tree.
As in our method, for every rule~(schema) application they define which child becomes the head
and impose a soft constraint that these dependencies agree with
the output of the dependency parser.
Our method is different in that we do not use the one-best dependency structure alone,
but rather we search for a CCG tree that is optimal in terms of dependencies and CCG supertags.
\citet{N10-1090} use the syntactic dependencies in a different way,
and show that dependency-based features are useful for predicting HPSG supertags.

In the CCG parsing literature, some work optimizes a {\it dependency model},
instead of supertags or a derivation~\cite{J07-4004,P14-1021}.
This approach is reasonable given that
the objective matches the evaluation metric.
Instead of modeling dependencies alone, our method
finds a CCG derivation that has a higher dependency score.
\citet{D15-1169} present a joint model of CCG parsing and semantic role labeling (SRL),
which is closely related to our approach. They map each CCG semantic dependency to an SRL relation,
for which they give the A* upper bound by the score from a predicate to the most probable argument.
Our approach is similar; the largest difference is that we instead model syntactic dependencies from each token to its head,
and this is the key to our success. Since dependency parsing can be formulated as independent head selections similar to tagging,
we can build the entire model on LSTMs to exploit features from the whole sentence.
This formulation is not straightforward in the case of multi-headed semantic dependencies in their model.

\section{Conclusion}

We have presented a new A* CCG parsing method,
in which the probability of a CCG tree is decomposed into
local factors of the CCG categories and its dependency structure.
By explicitly modeling the dependency structure,
we do not require any deterministic heuristics to resolve
attachment ambiguities,
and keep the model locally factored so that
all the probabilities can be precomputed before running the search.
Our parser efficiently finds the optimal parse
and achieves the state-of-the-art performance in both
English and Japanese parsing.


\section*{Acknowledgments}
We are grateful to Mike Lewis for answering our questions
and your Github repository from which we learned many things.
We also thank Yuichiro Sawai for the faster LSTM implementation.
This work was in part supported by JSPS KAKENHI
Grant Number 16H06981, and also by JST CREST Grant Number JPMJCR1301.

\bibliography{ccg}

\begin{thebibliography}{}
\expandafter\ifx\csname natexlab\endcsname\relax\def\natexlab#1{#1}\fi

\bibitem[{Abadi et~al.(2015)Abadi, Agarwal, Barham, Brevdo, Chen, Citro,
  Corrado, Davis, Dean, Devin, Ghemawat, Goodfellow, Harp, Irving, Isard, Jia,
  Jozefowicz, Kaiser, Kudlur, Levenberg, Man\'{e}, Monga, Moore, Murray, Olah,
  Schuster, Shlens, Steiner, Sutskever, Talwar, Tucker, Vanhoucke, Vasudevan,
  Vi\'{e}gas, Vinyals, Warden, Wattenberg, Wicke, Yu, and
  Zheng}]{tensorflow2015-whitepaper}
Mart\'{\i}n Abadi, Ashish Agarwal, Paul Barham, Eugene Brevdo, Zhifeng Chen,
  Craig Citro, Greg~S. Corrado, Andy Davis, Jeffrey Dean, Matthieu Devin,
  Sanjay Ghemawat, Ian Goodfellow, Andrew Harp, Geoffrey Irving, Michael Isard,
  Yangqing Jia, Rafal Jozefowicz, Lukasz Kaiser, Manjunath Kudlur, Josh
  Levenberg, Dan Man\'{e}, Rajat Monga, Sherry Moore, Derek Murray, Chris Olah,
  Mike Schuster, Jonathon Shlens, Benoit Steiner, Ilya Sutskever, Kunal Talwar,
  Paul Tucker, Vincent Vanhoucke, Vijay Vasudevan, Fernanda Vi\'{e}gas, Oriol
  Vinyals, Pete Warden, Martin Wattenberg, Martin Wicke, Yuan Yu, and Xiaoqiang
  Zheng. 2015.
\newblock \href{http://tensorflow.org/}{{TensorFlow: Large-Scale Machine
  Learning on Heterogeneous Systems}}.
\newblock Software available from tensorflow.org.
\newblock \href{http://tensorflow.org/}{http://tensorflow.org/}.

\bibitem[{Bangalore and Joshi(1999)}]{bangalore1999supertagging}
Srinivas Bangalore and Aravind~K Joshi. 1999.
\newblock {Supertagging: An Approach to Almost Parsing}.
\newblock {\em Computational linguistics\/} 25(2):237--265.

\bibitem[{Clark and Curran(2007)}]{J07-4004}
Stephen Clark and James~R. Curran. 2007.
\newblock \href{http://aclweb.org/anthology/J07-4004}{{Wide-Coverage Efficient
  Statistical Parsing with CCG and Log-Linear Models}}.
\newblock {\em Computational Linguistics, Volume 33, Number 4, December 2007\/}
  \href{http://aclweb.org/anthology/J07-4004}{http://aclweb.org/anthology/J07-4004}.

\bibitem[{Clevert et~al.(2015)Clevert, Unterthiner, and
  Hochreiter}]{DBLP:journals/corr/ClevertUH15}
Djork{-}Arn{\'{e}} Clevert, Thomas Unterthiner, and Sepp Hochreiter. 2015.
\newblock \href{http://arxiv.org/abs/1511.07289}{{Fast and Accurate Deep
  Network Learning by Exponential Linear Units (ELUs)}}.
\newblock {\em CoRR\/} abs/1511.07289.
\newblock
  \href{http://arxiv.org/abs/1511.07289}{http://arxiv.org/abs/1511.07289}.

\bibitem[{dos Santos and Zadrozny(2014)}]{santos:ICML}
C\'icero~Nogueira dos Santos and Bianca Zadrozny. 2014.
\newblock {Learning Character-level Representations for Part-of-Speech
  Tagging}.
\newblock ICML.

\bibitem[{Dozat and Manning(2016)}]{DBLP:journals/corr/DozatM16}
Timothy Dozat and Christopher~D. Manning. 2016.
\newblock \href{http://arxiv.org/abs/1611.01734}{{Deep Biaffine Attention for
  Neural Dependency Parsing}}.
\newblock {\em CoRR\/} abs/1611.01734.
\newblock
  \href{http://arxiv.org/abs/1611.01734}{http://arxiv.org/abs/1611.01734}.

\bibitem[{Dyer et~al.(2015)Dyer, Ballesteros, Ling, Matthews, and
  Smith}]{P15-1033}
Chris Dyer, Miguel Ballesteros, Wang Ling, Austin Matthews, and A.~Noah Smith.
  2015.
\newblock \href{https://doi.org/10.3115/v1/P15-1033}{{Transition-Based
  Dependency Parsing with Stack Long Short-Term Memory}}.
\newblock In {\em Proceedings of the 53rd Annual Meeting of the Association for
  Computational Linguistics and the 7th International Joint Conference on
  Natural Language Processing (Volume 1: Long Papers)\/}. Association for
  Computational Linguistics, pages 334--343.
\newblock
  \href{https://doi.org/10.3115/v1/P15-1033}{https://doi.org/10.3115/v1/P15-1033}.

\bibitem[{Eisner(1996)}]{P96-1011}
Jason Eisner. 1996.
\newblock \href{http://aclweb.org/anthology/P96-1011}{{Efficient Normal-Form
  Parsing for Combinatory Categorial Grammar}}.
\newblock In {\em 34th Annual Meeting of the Association for Computational
  Linguistics\/}.
\newblock
  \href{http://aclweb.org/anthology/P96-1011}{http://aclweb.org/anthology/P96-1011}.

\bibitem[{Hockenmaier and Bisk(2010)}]{C10-1053}
Julia Hockenmaier and Yonatan Bisk. 2010.
\newblock \href{http://aclweb.org/anthology/C10-1053}{{Normal-form parsing for
  Combinatory Categorial Grammars with generalized composition and
  type-raising}}.
\newblock In {\em Proceedings of the 23rd International Conference on
  Computational Linguistics (Coling 2010)\/}. Coling 2010 Organizing Committee,
  pages 465--473.
\newblock
  \href{http://aclweb.org/anthology/C10-1053}{http://aclweb.org/anthology/C10-1053}.

\bibitem[{Hockenmaier and Steedman(2007)}]{ccgbank}
Julia Hockenmaier and Mark Steedman. 2007.
\newblock \href{http://www.aclweb.org/anthology/J07-3004}{{CCGbank: A Corpus of
  CCG Derivations and Dependency Structures Extracted from the Penn Treebank}}.
\newblock {\em Computational Linguistics\/} 33(3):355--396.
\newblock
  \href{http://www.aclweb.org/anthology/J07-3004}{http://www.aclweb.org/anthology/J07-3004}.

\bibitem[{Kiperwasser and Goldberg(2016)}]{TACL885}
Eliyahu Kiperwasser and Yoav Goldberg. 2016.
\newblock
  \href{https://www.transacl.org/ojs/index.php/tacl/article/view/885}{{Simple
  and Accurate Dependency Parsing Using Bidirectional LSTM Feature
  Representations}}.
\newblock {\em Transactions of the Association for Computational Linguistics\/}
  4:313--327.
\newblock
  \href{https://www.transacl.org/ojs/index.php/tacl/article/view/885}{https://www.transacl.org/ojs/index.php/tacl/article/view/885}.

\bibitem[{Klein and D.~Manning(2003)}]{klein:a_star}
Dan Klein and Christopher D.~Manning. 2003.
\newblock \href{http://aclweb.org/anthology/N03-1016}{{A* Parsing: Fast Exact
  Viterbi Parse Selection}}.
\newblock In {\em Proceedings of the 2003 Human Language Technology Conference
  of the North American Chapter of the Association for Computational
  Linguistics\/}.
\newblock
  \href{http://aclweb.org/anthology/N03-1016}{http://aclweb.org/anthology/N03-1016}.

\bibitem[{Kudo and Matsumoto(2002)}]{DBLP:conf/conll/KudoM02}
Taku Kudo and Yuji Matsumoto. 2002.
\newblock \href{http://aclweb.org/anthology/W/W02/W02-2016.pdf}{{Japanese
  Dependency Analysis using Cascaded Chunking}}.
\newblock In {\em Proceedings of the 6th Conference on Natural Language
  Learning, CoNLL 2002, Held in cooperation with {COLING} 2002, Taipei, Taiwan,
  2002\/}.
\newblock
  \href{http://aclweb.org/anthology/W/W02/W02-2016.pdf}{http://aclweb.org/anthology/W/W02/W02-2016.pdf}.

\bibitem[{Lee et~al.(2016)Lee, Lewis, and
  Zettlemoyer}]{lee-lewis-zettlemoyer:2016:EMNLP2016}
Kenton Lee, Mike Lewis, and Luke Zettlemoyer. 2016.
\newblock \href{http://aclweb.org/anthology/D16-1262}{{Global Neural CCG
  Parsing with Optimality Guarantees}}.
\newblock In {\em Proceedings of the 2016 Conference on Empirical Methods in
  Natural Language Processing\/}. Association for Computational Linguistics,
  pages 2366--2376.
\newblock
  \href{http://aclweb.org/anthology/D16-1262}{http://aclweb.org/anthology/D16-1262}.

\bibitem[{Lei et~al.(2014)Lei, Xin, Zhang, Barzilay, and Jaakkola}]{P14-1130}
Tao Lei, Yu~Xin, Yuan Zhang, Regina Barzilay, and Tommi Jaakkola. 2014.
\newblock \href{https://doi.org/10.3115/v1/P14-1130}{{Low-Rank Tensors for
  Scoring Dependency Structures}}.
\newblock In {\em Proceedings of the 52nd Annual Meeting of the Association for
  Computational Linguistics (Volume 1: Long Papers)\/}. Association for
  Computational Linguistics, pages 1381--1391.
\newblock
  \href{https://doi.org/10.3115/v1/P14-1130}{https://doi.org/10.3115/v1/P14-1130}.

\bibitem[{Lewis et~al.(2015)Lewis, He, and Zettlemoyer}]{D15-1169}
Mike Lewis, Luheng He, and Luke Zettlemoyer. 2015.
\newblock \href{https://doi.org/10.18653/v1/D15-1169}{{Joint A* CCG Parsing and
  Semantic Role Labelling}}.
\newblock In {\em Proceedings of the 2015 Conference on Empirical Methods in
  Natural Language Processing\/}. Association for Computational Linguistics,
  pages 1444--1454.
\newblock
  \href{https://doi.org/10.18653/v1/D15-1169}{https://doi.org/10.18653/v1/D15-1169}.

\bibitem[{Lewis et~al.(2016)Lewis, Lee, and
  Zettlemoyer}]{lewis-lee-zettlemoyer:2016:N16-1}
Mike Lewis, Kenton Lee, and Luke Zettlemoyer. 2016.
\newblock \href{https://doi.org/10.18653/v1/N16-1026}{{LSTM CCG Parsing}}.
\newblock In {\em Proceedings of the 2016 Conference of the North American
  Chapter of the Association for Computational Linguistics: Human Language
  Technologies\/}. Association for Computational Linguistics, pages 221--231.
\newblock
  \href{https://doi.org/10.18653/v1/N16-1026}{https://doi.org/10.18653/v1/N16-1026}.

\bibitem[{Lewis and Steedman(2014)}]{lewis-steedman:2014:EMNLP2014}
Mike Lewis and Mark Steedman. 2014.
\newblock \href{https://doi.org/10.3115/v1/D14-1107}{{A* CCG Parsing with a
  Supertag-factored Model}}.
\newblock In {\em Proceedings of the 2014 Conference on Empirical Methods in
  Natural Language Processing (EMNLP)\/}. Association for Computational
  Linguistics, pages 990--1000.
\newblock
  \href{https://doi.org/10.3115/v1/D14-1107}{https://doi.org/10.3115/v1/D14-1107}.

\bibitem[{Neubig et~al.(2017)Neubig, Dyer, Goldberg, Matthews, Ammar,
  Anastasopoulos, Ballesteros, Chiang, Clothiaux, Cohn, Duh, Faruqui, Gan,
  Garrette, Ji, Kong, Kuncoro, Kumar, Malaviya, Michel, Oda, Richardson,
  Saphra, Swayamdipta, and Yin}]{dynet}
Graham Neubig, Chris Dyer, Yoav Goldberg, Austin Matthews, Waleed Ammar,
  Antonios Anastasopoulos, Miguel Ballesteros, David Chiang, Daniel Clothiaux,
  Trevor Cohn, Kevin Duh, Manaal Faruqui, Cynthia Gan, Dan Garrette, Yangfeng
  Ji, Lingpeng Kong, Adhiguna Kuncoro, Gaurav Kumar, Chaitanya Malaviya, Paul
  Michel, Yusuke Oda, Matthew Richardson, Naomi Saphra, Swabha Swayamdipta, and
  Pengcheng Yin. 2017.
\newblock {DyNet: The Dynamic Neural Network Toolkit}.
\newblock {\em arXiv preprint arXiv:1701.03980\/} .

\bibitem[{Noji and Miyao(2016)}]{noji-miyao:2016:P16-4}
Hiroshi Noji and Yusuke Miyao. 2016.
\newblock \href{https://doi.org/10.18653/v1/P16-4018}{{Jigg: A Framework for an
  Easy Natural Language Processing Pipeline}}.
\newblock In {\em Proceedings of ACL-2016 System Demonstrations\/}. Association
  for Computational Linguistics, pages 103--108.
\newblock
  \href{https://doi.org/10.18653/v1/P16-4018}{https://doi.org/10.18653/v1/P16-4018}.

\bibitem[{Pennington et~al.(2014)Pennington, Socher, and
  Manning}]{pennington2014glove}
Jeffrey Pennington, Richard Socher, and Christopher~D. Manning. 2014.
\newblock \href{http://www.aclweb.org/anthology/D14-1162}{{GloVe: Global
  Vectors for Word Representation}}.
\newblock In {\em Empirical Methods in Natural Language Processing (EMNLP)\/}.
  pages 1532--1543.
\newblock
  \href{http://www.aclweb.org/anthology/D14-1162}{http://www.aclweb.org/anthology/D14-1162}.

\bibitem[{Pollard and Sag(1994)}]{pollard1994head}
Carl Pollard and Ivan~A Sag. 1994.
\newblock {\em Head-driven phrase structure grammar\/}.
\newblock University of Chicago Press.

\bibitem[{Sagae et~al.(2007)Sagae, Miyao, and Tsujii}]{P07-1079}
Kenji Sagae, Yusuke Miyao, and Jun'ichi Tsujii. 2007.
\newblock \href{http://aclweb.org/anthology/P07-1079}{{HPSG Parsing with
  Shallow Dependency Constraints}}.
\newblock In {\em Proceedings of the 45th Annual Meeting of the Association of
  Computational Linguistics\/}. Association for Computational Linguistics,
  pages 624--631.
\newblock
  \href{http://aclweb.org/anthology/P07-1079}{http://aclweb.org/anthology/P07-1079}.

\bibitem[{Steedman(2000)}]{steedman:syntactic_process}
Mark Steedman. 2000.
\newblock {\em {The Syntactic Process}\/}.
\newblock The MIT Press.

\bibitem[{Tokui et~al.(2015)Tokui, Oono, Hido, and
  Clayton}]{chainer_learningsys2015}
Seiya Tokui, Kenta Oono, Shohei Hido, and Justin Clayton. 2015.
\newblock
  \href{http://learningsys.org/papers/LearningSys\_2015\_paper\_33.pdf}{{Chainer:
  a Next-Generation Open Source Framework for Deep Learning}}.
\newblock In {\em Proceedings of Workshop on Machine Learning Systems
  (LearningSys) in The Twenty-ninth Annual Conference on Neural Information
  Processing Systems (NIPS)\/}.
\newblock
  \href{http://learningsys.org/papers/LearningSys\_2015\_paper\_33.pdf}{http://learningsys.org/papers/LearningSys\_2015\_paper\_33.pdf}.

\bibitem[{Toutanova et~al.(2003)Toutanova, Klein, Manning, and
  Singer}]{N03-1033}
Kristina Toutanova, Dan Klein, Christopher~D. Manning, and Yoram Singer. 2003.
\newblock \href{http://www.aclweb.org/anthology/N03-1033}{{Feature-Rich
  Part-of-Speech Tagging with a Cyclic Dependency Network}}.
\newblock In {\em Proceedings of the 2003 Human Language Technology Conference
  of the North American Chapter of the Association for Computational
  Linguistics\/}.
\newblock
  \href{http://www.aclweb.org/anthology/N03-1033}{http://www.aclweb.org/anthology/N03-1033}.

\bibitem[{Turian et~al.(2010)Turian, Ratinov, and Bengio}]{P10-1040}
Joseph Turian, Lev-Arie Ratinov, and Yoshua Bengio. 2010.
\newblock \href{http://aclweb.org/anthology/P10-1040}{{Word Representations: A
  Simple and General Method for Semi-Supervised Learning}}.
\newblock In {\em Proceedings of the 48th Annual Meeting of the Association for
  Computational Linguistics\/}. Association for Computational Linguistics,
  pages 384--394.
\newblock
  \href{http://aclweb.org/anthology/P10-1040}{http://aclweb.org/anthology/P10-1040}.

\bibitem[{Uchimoto et~al.(1999)Uchimoto, Sekine, and Isahara}]{E99-1026}
Kiyotaka Uchimoto, Satoshi Sekine, and Hitoshi Isahara. 1999.
\newblock \href{http://aclweb.org/anthology/E99-1026}{{Japanese Dependency
  Structure Analysis Based on Maximum Entropy Models}}.
\newblock In {\em Ninth Conference of the European Chapter of the Association
  for Computational Linguistics\/}.
\newblock
  \href{http://aclweb.org/anthology/E99-1026}{http://aclweb.org/anthology/E99-1026}.

\bibitem[{Uematsu et~al.(2013)Uematsu, Matsuzaki, Hanaoka, Miyao, and
  Mima}]{uematsu-EtAl:2013:ACL2013}
Sumire Uematsu, Takuya Matsuzaki, Hiroki Hanaoka, Yusuke Miyao, and Hideki
  Mima. 2013.
\newblock \href{http://www.aclweb.org/anthology/P13-1103}{{Integrating Multiple
  Dependency Corpora for Inducing Wide-coverage Japanese CCG Resources}}.
\newblock In {\em Proceedings of the 51st Annual Meeting of the Association for
  Computational Linguistics (Volume 1: Long Papers)\/}. Association for
  Computational Linguistics, Sofia, Bulgaria, pages 1042--1051.
\newblock
  \href{http://www.aclweb.org/anthology/P13-1103}{http://www.aclweb.org/anthology/P13-1103}.

\bibitem[{Vaswani et~al.(2016)Vaswani, Bisk, Sagae, and Musa}]{N16-1027}
Ashish Vaswani, Yonatan Bisk, Kenji Sagae, and Ryan Musa. 2016.
\newblock \href{https://doi.org/10.18653/v1/N16-1027}{{Supertagging With
  LSTMs}}.
\newblock In {\em Proceedings of the 2016 Conference of the North American
  Chapter of the Association for Computational Linguistics: Human Language
  Technologies\/}. Association for Computational Linguistics, pages 232--237.
\newblock
  \href{https://doi.org/10.18653/v1/N16-1027}{https://doi.org/10.18653/v1/N16-1027}.

\bibitem[{Weiss et~al.(2015)Weiss, Alberti, Collins, and
  Petrov}]{weiss-EtAl:2015:ACL-IJCNLP}
David Weiss, Chris Alberti, Michael Collins, and Slav Petrov. 2015.
\newblock \href{https://doi.org/10.3115/v1/P15-1032}{{Structured Training for
  Neural Network Transition-Based Parsing}}.
\newblock In {\em Proceedings of the 53rd Annual Meeting of the Association for
  Computational Linguistics and the 7th International Joint Conference on
  Natural Language Processing (Volume 1: Long Papers)\/}. Association for
  Computational Linguistics, pages 323--333.
\newblock
  \href{https://doi.org/10.3115/v1/P15-1032}{https://doi.org/10.3115/v1/P15-1032}.

\bibitem[{Xu et~al.(2014)Xu, Clark, and Zhang}]{P14-1021}
Wenduan Xu, Stephen Clark, and Yue Zhang. 2014.
\newblock \href{https://doi.org/10.3115/v1/P14-1021}{{Shift-Reduce CCG Parsing
  with a Dependency Model}}.
\newblock In {\em Proceedings of the 52nd Annual Meeting of the Association for
  Computational Linguistics (Volume 1: Long Papers)\/}. Association for
  Computational Linguistics, pages 218--227.
\newblock
  \href{https://doi.org/10.3115/v1/P14-1021}{https://doi.org/10.3115/v1/P14-1021}.

\bibitem[{Zhang et~al.(2010)Zhang, Matsuzaki, and Tsujii}]{N10-1090}
Yao-zhong Zhang, Takuya Matsuzaki, and Jun'ichi Tsujii. 2010.
\newblock \href{http://aclweb.org/anthology/N10-1090}{{A Simple Approach for
  HPSG Supertagging Using Dependency Information}}.
\newblock In {\em Human Language Technologies: The 2010 Annual Conference of
  the North American Chapter of the Association for Computational
  Linguistics\/}. Association for Computational Linguistics, pages 645--648.
\newblock
  \href{http://aclweb.org/anthology/N10-1090}{http://aclweb.org/anthology/N10-1090}.

\end{thebibliography}
\bibliographystyle{acl_natbib}

\end{document}